\documentclass[10pt,twocolumn,letterpaper]{article}

\usepackage[pagenumbers]{cvpr} 
\usepackage{caption}
\usepackage{subcaption}
\usepackage{booktabs}
\usepackage{pifont}
\usepackage{multirow}
\usepackage{graphicx}
\usepackage{algorithm}
\usepackage{algpseudocode}
\usepackage{float}
\usepackage{tabularx}
\definecolor{cvprblue}{rgb}{0.21,0.49,0.74}
\usepackage[pagebackref,breaklinks,colorlinks,allcolors=cvprblue]{hyperref}

\def\paperID{9781} 
\def\confName{CVPR}
\def\confYear{2025}

\title{FedCCRL: Federated Domain Generalization with Cross-Client Representation Learning}

\author{
    Xinpeng Wang \quad Yongxin Guo \quad Xiaoying Tang\\  
    The Chinese University of Hong Kong, Shenzhen\\  
    {\tt\small {xinpengwang@link.cuhk.edu.cn} \quad {yongxinguo@link.cuhk.edu.cn} \quad {tangxiaoying@cuhk.edu.cn}}  
}

\begin{document}
\maketitle
\begin{abstract}
Domain Generalization (DG) aims to train models that can effectively generalize to unseen domains. However, in the context of Federated Learning (FL), where clients collaboratively train a model without directly sharing their data, most existing DG algorithms are not directly applicable to the FL setting due to privacy constraints, as well as the limited data quantity and domain diversity at each client. To tackle these challenges, we propose FedCCRL, a lightweight federated domain generalization method that significantly improves the model’s generalization ability while preserving privacy and ensuring computational and communication efficiency. Specifically, FedCCRL comprises two principal modules: the first is a cross-client feature extension module, which increases local domain diversity via cross-client domain transfer and domain-invariant feature perturbation; the second is a representation and prediction dual-stage alignment module, which enables the model to effectively capture domain-invariant features. 
Extensive experimental results demonstrate that FedCCRL achieves the state-of-the-art performance on the PACS, OfficeHome and miniDomainNet datasets across FL settings of varying numbers of clients. 
Code is available at 
\href{https://github.com/sanphouwang/fedccrl}{https://github.com/sanphouwang/fedccrl}.
\end{abstract}    
\section{Introduction}
\label{sec:intro}
Traditional machine learning algorithms typically assume that training and test data follow the same independent and identically distributed (IID) pattern. However, this assumption is difficult to maintain in real-world applications, where models often encounter Out-of-Distribution (OOD) data \cite{recht2019imagenet}. In such cases, models trained on specific source domains (\eg, cartoon images) tend to suffer significant performance degradation on unseen domains (\eg, sketch images). To address the challenges posed by domain shift, Domain Generalization (DG) \cite{DGsurvey}, which aims to enhance the model’s ability to generalize to unseen domains, has garnered significant attention. Nevertheless, traditional DG algorithms assume that the data is centrally stored and accessible to a single model, which contrasts with real-world scenarios where data is distributed across multiple clients. To enable these clients to collaboratively train a model without exchanging their raw data, Federated Learning (FL) \cite{fedavg} emerges as a promising framework.

\begin{figure}[t]
    \centering
    \includegraphics[width=0.9\linewidth]{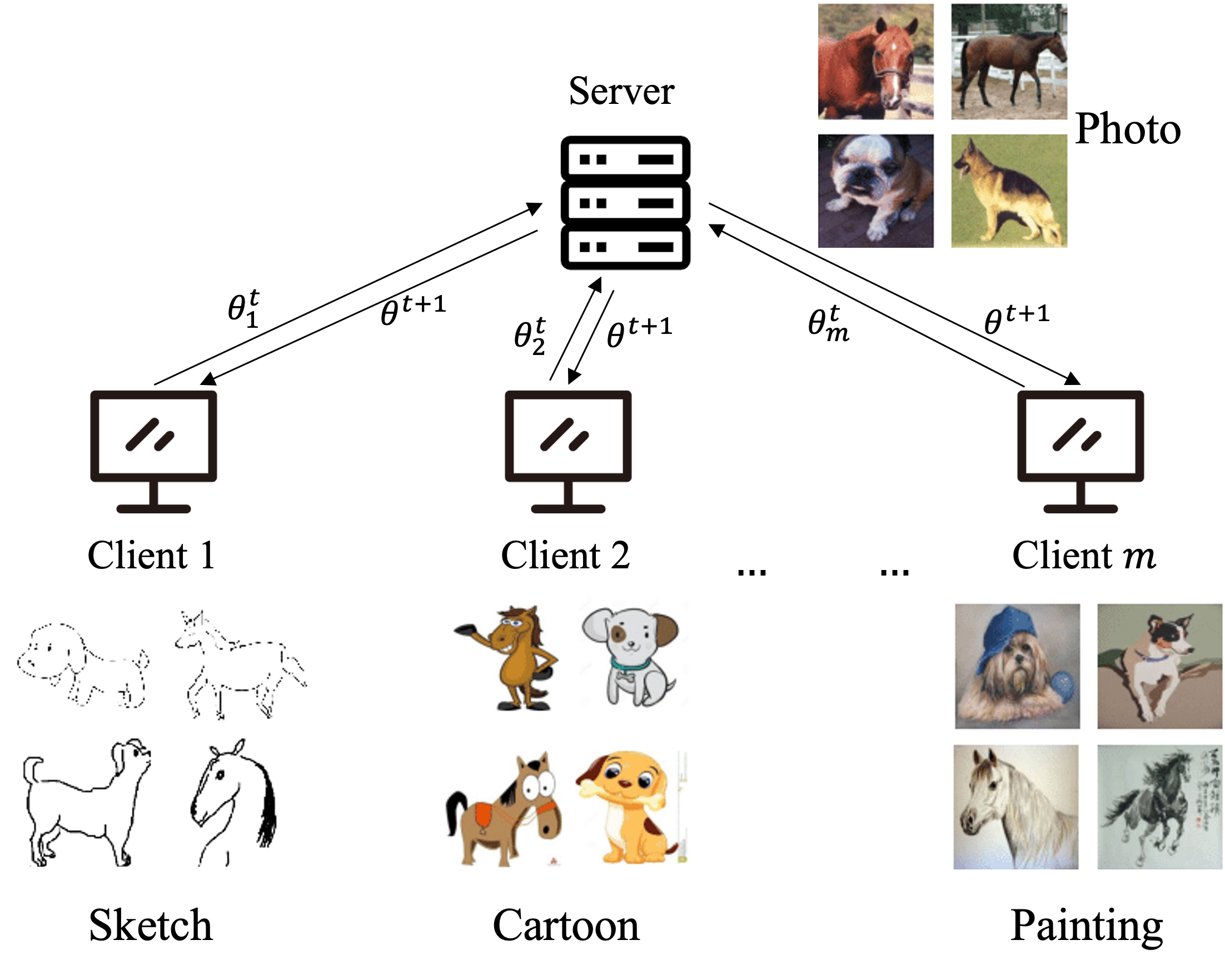}
    \caption{Illustration of the typical scenario in FL. Each client contains data from a unique domain, and the test domain (Photo) differs from all domains present on the clients.}
    \label{fig: federated domain generalization setting}
\end{figure}

However, most existing DG methods are not directly applicable to the FL setting due to two primary limitations. One key limitation is that traditional DG algorithms frequently depend on inter-domain data sharing, which contravenes the fundamental privacy principles of FL \cite{fl_privacy_survey1,fl_privacy_survey2}. Another significant challenge, as illustrated in \cref{fig: federated domain generalization setting}, is the limited data quantity and insufficient domain diversity within individual clients, which complicates achieving satisfactory performance when using conventional centralized DG approaches \cite{li2018domain,zhao2020domain}.

Most current FDG methods aim to enhance the model's generalization capability by learning domain-invariant features (\ie, label-relevant features that are consistent across domains), typically through federated adversarial learning \cite{fedadg,federated_ad_training_1,federated_ad_training_2, federated_ad_training_3} or federated representation alignment \cite{fedsr,federated_representation_alignment_1,federated_representation_alignment_2}. Although both approaches effectively assist models in capturing domain-invariant features, each has its own limitations. 
Federated adversarial learning introduces substantial computational overhead due to the use of a discriminator and encounters training instability because of the risk of model collapse \cite{model_collapse}. On the other hand, federated representation alignment methods, though more computationally efficient, experience performance constraints arising from limited data quantity and domain diversity on each client, as well as privacy-preserving restrictions that preclude direct data sharing.

Another major approach for FDG is federated style transfer \cite{yang2020fda,yoon2021fedmix}, which utilizes FL-compatible style transfer techniques, such as AdaIN \cite{AdaIN} and CycleGAN \cite{cyclegan}, to enhance domain diversity on each client. However, most of these methods impose significant computational and transmission overheads, as they involve not only additional models for feature extraction and sample reconstruction but also the transmission of high-dimensional feature embeddings \cite{stablefdg,ccst}. 
Moreover, these algorithms focus exclusively on transferring domain-specific features (\ie, features that vary across domains and are minimally related to the label) and ignore the augmentation for domain-invariant features, which limits the diversity of generated samples.

Given the limited domain diversity and the constraints on computational and transmission costs in the FL setting, a lightweight FDG framework is needed to enhance local domain diversity and enable the model to capture domain-invariant features effectively. To this end, we propose FedCCRL, a lightweight FDG method that significantly improves the model's generalization capability in the FL setting while maintaining privacy and incurring only negligible additional computational and communication costs. To mitigate the adverse effects of insufficient domain diversity on federated representation alignment, FedCCRL employs a lightweight and privacy-preserving cross-client feature extension module, which effectively transfers domain-specific features across domains and perturbs domain-invariant features. To further enable the model to extract domain-invariant features, FedCCRL applies a representation and prediction dual-stage alignment using supervised contrastive loss and Jensen-Shannon divergence.

The key contributions of this work are outlined as follows:
\begin{itemize}
    \item We propose FedCCRL, a lightweight FDG framework that effectively enhances domain diversity on each client and enables the model to capture domain-invariant features, without incurring substantial computational or transmission costs.

    \item Extensive experiments on PACS, OfficeHome and miniDomainNet datasets demonstrate that the proposed FedCCRL method achieves the state-of-the-art performance across FL systems of varying numbers of clients.
\end{itemize}
\section{Related Work}
\subsection{Representation Alignment}
A fundamental technique in representation learning-based DG methods is representation alignment, where feature representations from different domains are aligned to minimize domain-specific variations. For instance, DANN \cite{dann1,dann2,dann3} employs adversarial learning to align the representation distributions across domains by using a domain classifier to ensure that the learned features are domain-invariant. Another prominent approach is CORAL \cite{coral}, which aligns the second-order statistics of source and target representation distributions to reduce domain shift. Additionally, MMD-based methods \cite{mmd1,mmd2,mmd3} utilize kernel-based distances to align representations across domains. 

Although these methods help models generalize better to unseen domains, most of them require the access to all of the training domains, which violates privacy principles in FL as direct data transmission between clients or to a central server is strictly prohibited.
\subsection{Style Transfer}
Style transfer-based DG methods \cite{amvdsi,volpi2018generalizing,xu2021robust} enhance domain diversity to improve model robustness across unseen domains and can be divided into two main categories.

The first category leverages generative models to synthesize data with diverse styles \cite{robey2021model,palakkadavath2024domain}, thereby reducing the model's reliance on domain-specific features. Despite their effectiveness, these generative approaches entail substantial computational costs and are prone to issues like model collapse during adversarial training, which can hinder stable convergence.

The second category includes augmentation-based techniques, such as MixStyle \cite{mixstyle} and Mixup \cite{mixup}, which diversify training data without explicit generation. MixStyle interpolates style representations within a batch, promoting domain-invariant feature learning, while Mixup creates interpolated samples that encourage generalization across domain boundaries. These approaches are more computationally efficient and avoid the instability often associated with generative models, making them advantageous for scalable DG applications.

\subsection{Federated Domain Generalization}
Most existing FDG algorithms rely on domain-invariant representation learning through federated adversarial learning or federated representation alignment.
For instance, FedADG \cite{fedadg} employs a global discriminator to enable the model to extract universal feature representations across clients while holding privacy principles. In addition, FedSR \cite{fedsr} adopts an L2-norm regularizer and a conditional mutual information regularizer to align representation distributions across clients. However, federated adversarial learning entails significant computational overhead and the risk of model collapse. Federated representation alignment, on the other hand, demonstrates limited effectiveness in large-scale federated learning contexts due to insufficient domain diversity on individual clients.

While these FDG algorithms have contributed to enhancing the model's generalization ability to a certain extent, they remain insufficient in effectively addressing the constraints of limited data volume and domain diversity on each client. To this end,  CCST \cite{ccst} employs a cross-client style transfer approach based on AdaIN \cite{AdaIN} to generate samples with styles from other domains on the client side. 
However, CCST relies on a pre-trained VGG network \cite{vgg} to extract feature representations and reconstruct samples, while also needing to transmit these high-dimensional representations, which results in significant additional communication and computational overheads. Furthermore, transmitting these feature representations violates privacy protection principles in FL, as attackers can easily intercept these representations and reconstruct the samples with the same decoder \cite{fl_privacy_survey1,fl_privacy_survey2}. Additionally, the use of a pre-trained VGG network also partially violates the principles of domain generalization, as the target domain might be included in the pre-training dataset.

Apart from the methods discussed above, other FDG approaches take different strategies. For instance, FedIIR \cite{fediir} implicitly learns invariant relationships across domains by aligning gradients, which allows the model to generalize to OOD data. Besides, GA \cite{ga} aids FDG by dynamically calibrating aggregation weights, thus minimizing generalization gaps across clients. 
However, these methods have not demonstrated significant advantages over other FDG approaches in experimental performance.

\section{Method}
\begin{figure*}[t]
    \centering
    \includegraphics[width=\textwidth]{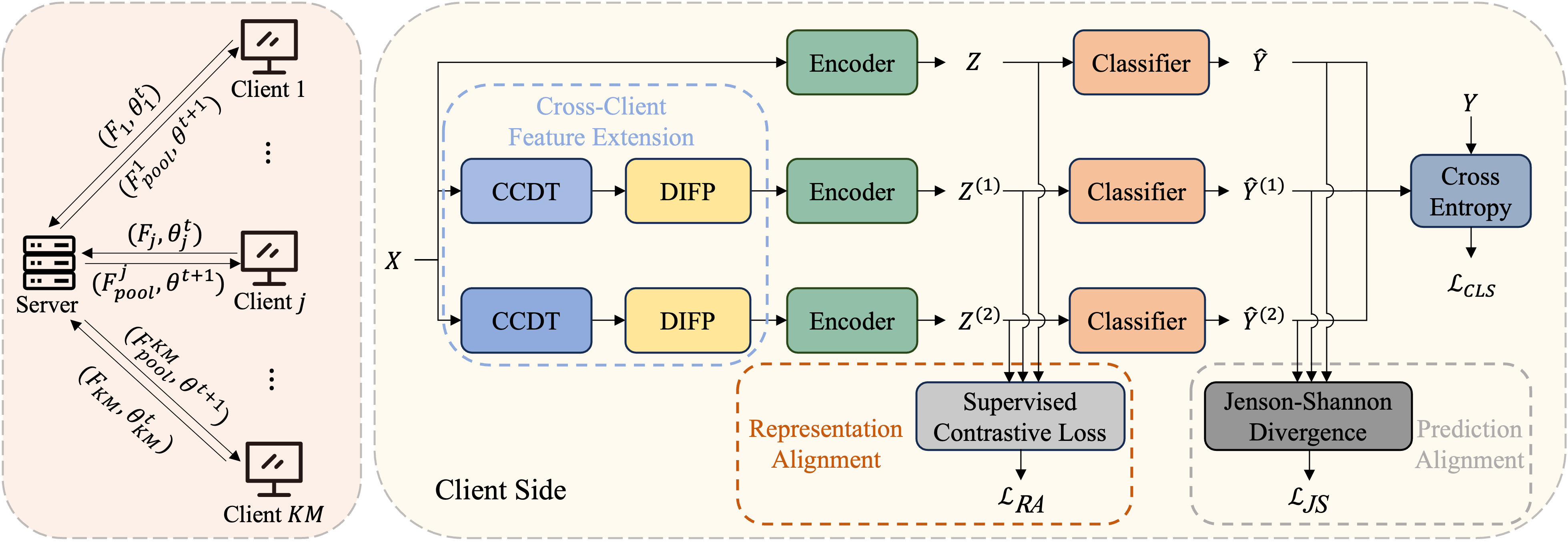}
    \caption{Overview of FedCCRL. The client transmits local model parameters and sample statistics to the server, where these elements are aggregated and redistributed to each client. During the local training phase, feature augmentation, representation alignment and prediction alignment are applied to ensure the model learns to focus on domain-invariant features. }
    \label{fig:overview}
\end{figure*}

\subsection{Preliminary}
\label{sec: preliminary}
Let $\mathcal{X}$ and $\mathcal{Y}$ denote the input space and target space respectively. In FDG, given $M$ source domains $S_{source}=\{S_i | i=1,2,\dots,M\}$, the data points in $S_i$ are independently and identically drawn from distribution $P_i(x, y)$, where $x \in \mathcal{X}$, $y \in \mathcal{Y}$.  The joint distribution between each pair of domains are different, \ie, $P_i(x, y) \neq P_j(x, y)$ for $1 \leq i \neq j \leq M$. 


In this study, we adopt a commonly used data partitioning scheme in FDG, where the dataset on each client, denoted as $D_i$, originates exclusively from a single domain. Specifically, each source domain is partitioned into $K$ clients: $\cup_{j=k\cdot(i-1)+1}^{i\cdot k} D_j=S_i$, and $D_q \cap D_p =\emptyset$ for $1\leq q \neq p \leq K\cdot M$.

The goal of FDG is to enable these $K \cdot M$ clients to collaboratively train a model $f: \mathcal{X} \to \mathcal{Y}$, such that the model is capable of generalizing well and minimize the prediction error on a target domain $S_{\text{target}}$:
\begin{equation}
    \min_{f} \mathbb{E}_{(x, y) \sim S_{\text{target}}} \left[ l(f(x), y) \right],
\end{equation}
where $l(\cdot)$ represents the loss function. Samples in $S_{\text{target}}$ are drawn from $P_{\text{target}}$ and $P_{\text{target}}(x, y) \neq P_i(x, y)$ for $i \in \{1, 2, \dots, M\}$.

To protect privacy, the server is not permitted to directly access the data on the clients. After local training, each client uploads its model parameters to the server, where the parameters are aggregated and then sent back to the clients. Let $\theta_i^t$ denote the model parameters of client $i$ at round $t$. The server aggregates the parameters as follows:
\begin{equation}
    \theta^{t+1} = \frac{1}{N} \sum_{i=1}^{K \cdot M} n_i \theta_i^t,    
    \label{eq: agg model}
\end{equation}
where $N = \sum_{i=1}^{K \cdot M} n_i$ is the total number of samples across all clients \cite{fedavg} and $n_i$ denotes the number of samples on client $i$. After aggregation, the server sends the updated parameters $\theta^{t+1}$ back to each client.

\subsection{Algorithm Overview}
We provide the overview of the proposed FedCCRL framework in \cref{fig:overview}, with the algorithmic details outlined in \cref{alg: FedCCRL}. 

On the client side, for a batch of samples $X \in \mathbb{R}^{B \times C \times H \times W}$, where $B$ is the batch size, $C$ is the number of channels, and $H$ and $W$ represent the height and width of the images respectively, we first apply cross-client feature extension to generate two batches of augmented samples:
\begin{equation}
    X^{(1)}=\mathcal{M}(X), X^{(2)}=\mathcal{M}(X),
    \label{eq: feature augmentation}
\end{equation}
where $\mathcal{M}(\cdot)=\text{CCDT}\circ \text{DIFP}$ represents the composition of CCDT and DIFP in the cross-client feature extension module. The detailed procedures for these components are provided in \cref{alg: CCDT} and \cref{alg: augmix} respectively.

After cross-client feature extension, we extract representations and generate predictions for each batch, which are then used to compute the overall loss. Let $\mathcal{Z}$ denote the representation space. We formalize the model $f$ as $f = g \circ h$, where $h: \mathcal{X} \rightarrow \mathcal{Z}$ represents the representation encoder and $g: \mathcal{Z} \rightarrow \hat{\mathcal{Y}}$ denotes the classifier. We extract representations and produce predictions as follows: 
\begin{equation}
    Z = h(X), Z^{(1)} = h(X^{(1)}), Z^{(2)} = h(X^{(2)}),
    \label{eq: representation extraction}
\end{equation}
\begin{equation}
    \hat{Y} = g(Z), \hat{Y}^{(1)} = g(Z^{(1)}), \hat{Y}^{(2)} = g(Z^{(2)}).
    \label{eq: prediction generation}
\end{equation}

Subsequently, we utilize supervised contrastive loss for representation alignment, as defined in \cref{eq: feature alignment loss}, and employ Jensen-Shannon divergence for prediction alignment, as specified in \cref{eq: js divergence}, to ensure consistent predictions between the original and generated samples.
\begin{algorithm}
    \caption{FedCCRL}
    \label{alg: FedCCRL}
    \textbf{Input:} Ratio of statistics to be uploaded $r\in(0,1)$, Hyper-parameters $\lambda_1$ and $\lambda_2$, Communication rounds $T$, Training epochs $E$.\\
    \textbf{Output:} Model parameter $\theta$
    \begin{algorithmic}[1]
    \State Server initializes $f$ and distributes it to each client.
    \For{each round $t = 1,2,\ldots,T$}
        \For{each client $i = 1,2,\ldots,KM$}
            \State $F_i\leftarrow$ \cref{eq: F_i}
            \State Client $i$ uploads $F_i$ to the server 
        \EndFor
        \State Server constructs $F_{pool}$ and distributes $F_{pool}^i$ to each client $i$
        \For{each client $i = 1,2,\ldots,KM$}
            \For{each epoch $e = 1,2,\ldots,E$}
                \For{each batch $X$ in $D_i$}
                    \State $X^{(1)}, X^{(2)}$$\leftarrow$ \cref{eq: feature augmentation}
                    \State $Z, Z^{(1)}, Z^{(2)}$$\leftarrow$ \cref{eq: representation extraction}
                    \State $\hat{Y}, \hat{Y}^{(1)}, \hat{Y}^{(2)}$$\leftarrow$ \cref{eq: prediction generation}
                    \State $\mathcal{L}_{RA}\leftarrow$ \cref{eq: feature alignment loss}
                    \State $\mathcal{L}_{JS}\leftarrow$ \cref{eq: js  divergence}
                    \State $\mathcal{L}_{CLS}\leftarrow$ \cref{eq: L_CLS}
                    \State $\mathcal{L}\leftarrow$ \cref{eq: overall_loss}
                    \State $\theta_i^t\leftarrow$ update model with Adam
                \EndFor
            \EndFor
            \State Client $i$ sends $\theta_i^t$ to the server
        \EndFor
        \State $\theta^{t+1}\leftarrow$ \cref{eq: agg model}
    \EndFor
    \State \textbf{return} $\theta^{T+1}$
    \end{algorithmic}
\end{algorithm}

\subsection{Lightweight Cross-Client Feature Extension}
While multiple studies \cite{bui2021exploiting,li2021simple} have shown that combining domain-specific and domain-invariant feature augmentations can create a more adaptable feature space, most data augmentation-based FDG methods transfer only domain-specific features and often introduce significant computational and transmission overhead. 

In contrast, FedCCRL introduces a lightweight cross-client feature extension module comprising two components: Cross-Client Domain Transfer (CCDT) and Domain-Invariant Feature Perturbation (DIFP). 
This cross-client feature extension module effectively transfer samples' domain-specific features across clients and perturbs domain-invariant features without incurring excessive computational and transmission costs.

CCDT, inspired by MixStyle \cite{mixstyle}, leverages sample statistics to transfer domain-specific features. Specifically, in FedCCRL, clients are required not only to transmit model parameters to the server but also to upload the channel-wise means and standard deviations of samples: 
\begin{equation}
    \mu(x)_c= \frac{1}{HW} \sum_{h=1}^{H}\sum_{w=1}^{W} x_{c,h,w}\text{,}
    \label{eq: mean}
\end{equation}
\begin{equation}
    \sigma(x)_c= \sqrt{\frac{1}{HW} \sum_{h=1}^{H}\sum_{w=1}^{W} (x_{c,h,w} - \mu(x)_c)^2}\text{,}
    \label{eq: std}
\end{equation}
where $x \in \mathbb{R}^{C \times H \times W}$ is a given sample.


Each client $i$ uploads a certain proportion of sample statistics denoted as:
\begin{equation}
    F_i=\{(\mu(x_j),\sigma(x_j)) \mid x_j \in D_i,j=1, 2, \ldots, \lceil rn_i \rceil\},
    \label{eq: F_i}
\end{equation} 
where $r \in (0, 1)$ represents the ratio of statistics to be uploaded. 

After collecting statistics from all clients, the server concatenates these statistics to get the pool of statistics: $ F_{\text{pool}} = \{F_i \mid i=1,2,\ldots,K\cdot M \} $, and subsequently distributes $F_{pool}^i=F_{pool} \setminus F_i$ to each client $i$. Each client apply CCDT to generate style transferred samples according to \cref{alg: CCDT}. 
\begin{algorithm}
    \caption{CCDT on Client $i$}
    \label{alg: CCDT}
    \textbf{Input:} A batch of samples $X$, Pool of statistics $F_{pool}^i$, $\alpha \in (0,\infty)$\\
    \textbf{Output:} A batch of style transferred samples $X_{CCDT}$
    \begin{algorithmic}[1]
    \State $\hat{X}=[ \ ]$
    \For{$x$ in $X$}
        \State Compute $\mu(x)$, $\sigma(x)$ \Comment{\cref{eq: mean}, \cref{eq: std}} 
        \State Normalize the input sample: $\tilde{x}=\frac{x-\mu(x)}{\sigma(x)}$
        \State Uniformly sample a pair of statistics $(\mu^\prime,\sigma^\prime) \sim F_{pool}^i$
        \State Sample $\lambda \sim \text{Beta}(\alpha,\alpha)$
        \State $\gamma_{mix}=\lambda\mu^\prime+(1-\lambda)\mu(x)$
        \State $\beta_{mix}=\lambda\sigma^\prime+(1-\lambda)\sigma(x)$
        \State $\hat{x}=\gamma_{mix}\tilde{x}+\beta_{mix}$
        \State $\hat{X}$.append($\hat{x}$)
    \EndFor
    \State $X_{CCDT}=\text{cat}(\hat{X})$
    \end{algorithmic}
\end{algorithm}

In DIFP, we employ AugMix \cite{augmix} to perturb domain-invariant features. AugMix takes the original sample $x$ and processes it through multiple augmentation chains, each consisting of a sequence of simple augmentation operations. The outputs from these chains are then mixed using an element-wise convex combination. The detailed process of AugMix is provided in \cref{alg: augmix} in supplementary material, and we adopt the same set of operations $\mathcal{O}$ as described in \cite{augmix}.

\begin{figure}[tbp]
    \centering
    \begin{subfigure}{0.3\linewidth}  
        \centering
        \includegraphics[width=\linewidth]{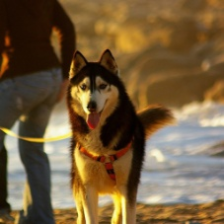}
        \caption{Original Image}
    \end{subfigure}
    \hfill
    \begin{subfigure}{0.3\linewidth}
        \centering
        \includegraphics[width=\linewidth]{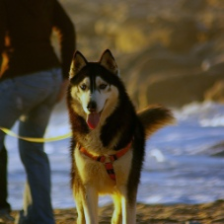}
        \caption{CCDT}
    \end{subfigure}
    \hfill
    \begin{subfigure}{0.3\linewidth}
        \centering
        \includegraphics[width=\linewidth]{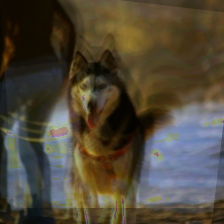}
        \caption{DIFP}
    \end{subfigure}
    \caption{Effect of feature augmentation. (b) is generated from (a) by CCDT, and (c) is generated from (b) by DIFP.}
    \label{fig: augmentation effect}
\end{figure}

In \cref{fig: augmentation effect}, we present the effects of CCDT and DIFP. It is evident that CCDT transfers domain-specific features, such as color distribution, while DIFP effectively perturbs domain-invariant features, such as contours of objects.

\subsection{Representation Alignment}
To enable the model to focus on extracting domain-invariant features that capture the essential information relevant to the task, we employ supervised contrastive loss \cite{supervised_contrastive_loss} to align feature representations across original samples and augmented samples. Supervised contrastive loss leverages label information to ensure that representations from the same class are pulled closer together in the latent space, while representations from different classes are pushed apart. This alignment mechanism enables the model to extract correct semantic features.

Given any two batches of representations $Z^\prime,Z^{\prime\prime}\in \mathbb{R}^{B\times V}$ and their corresponding labels $Y^\prime$ and $Y^{\prime\prime}$, we concatenate these two batches of features and labels:
\begin{equation}
    Z=\text{cat}(Z^\prime,Z^{\prime\prime})\in \mathbb{R}^{2B\times V},
\end{equation}
\begin{equation}
     Y=\text{cat}(Y^\prime,Y^{\prime\prime}),
\end{equation}
where $V$ denotes the size of feature representations.

Let $I=\{1,2,\ldots,2B\}$ to be the index set and $A(i)=I \setminus i$. The supervised contrastive loss of $Z^\prime$ and $Z^{\prime\prime}$ is calculated as follow:

\begin{equation}
    \mathcal{L}_{SC}(Z^\prime,Z^{\prime\prime}) =\sum_{i\in I} \frac{-1}{|P(i)|}\sum_{p \in P(i)} \log \frac{s(Z_i, Z_p)}{\sum_{a \in A(i)}s(Z_i, Z_a)},
\end{equation}
where
$P(i)=\{p\in A(i) \mid y_p=y_i\}$, $s(Z_i,Z_p)=\exp(\text{sim}(Z_i, Z_p)/\tau)$. $\text{sim}(\cdot,\cdot)$ denotes cosine similarity and $\tau$ is a temperature parameter that controls the scale of similarities.

The representation alignment loss \( \mathcal{L}_{\text{RA}} \) in FedCCRL is defined as:
\begin{equation}
    \mathcal{L}_{RA} = \frac{1} {2}(\mathcal{L}_{SC}(Z^{(1)},Z)+\mathcal{L}_{SC}(Z^{(2)},Z)).
    \label{eq: feature alignment loss}
\end{equation}

\subsection{Prediction Alignment}
Given that \( X^{(1)} \) and \( X^{(2)} \) are generated from \( X \), we seek for the model to accurately classify both the original and augmented samples while also ensuring consistency in the predictions, represented by \( \hat{Y} \), \( \hat{Y}^{(1)} \), and \( \hat{Y}^{(2)} \). 

To promote consistent predictions across the original and generated samples, we employ Jensen-Shannon (JS) divergence minimization. The JS divergence is specifically constructed to measure similarity between probability distributions by calculating their average and then measuring each distribution's divergence from this mean distribution. Formally, let the average prediction distribution be defined as follows:
\begin{equation}
    \Bar{Y} = \frac{1}{3}(\hat{Y} + \hat{Y}^{(1)} + \hat{Y}^{(2)}),
\end{equation}

where $\Bar{Y}$ represents the mean distribution of the predictions. The JS divergence loss is then computed as:
\begin{equation}
     \mathcal{L}_{JS}=\frac{1}{3}(\text{KL}(\hat{Y},\Bar{Y})+\text{KL}(\hat{Y}^{(1)},\Bar{Y})+\text{KL}(\hat{Y}^{(2)},\Bar{Y}))
    \label{eq: js  divergence}
\end{equation}
where $\text{KL}(\cdot, \cdot)$ denotes KL divergence. Minimizing $\mathcal{L}_{\text{JS}}$ aligns the model's predictions across original and augmented samples, thus enhancing prediction consistency.

%

\subsection{Overall Loss Function}
The overall loss function is formulated as follows:
\begin{equation} \mathcal{L}=\mathcal{L}_{CLS}+\lambda_1\mathcal{L}_{RA}+\lambda_2\mathcal{L}_{JS}, \label{eq: overall_loss} \end{equation}
where the classification loss is computed as:
\begin{equation} \mathcal{L}_{CLS}=\sum_{Y^\prime \in \{ \hat{Y},\hat{Y}^{(1)},\hat{Y}^{(2)}\}}\frac{1}{3}\mathcal{L}_{CRE}(Y^\prime,Y), \label{eq: L_CLS} \end{equation}
and $\mathcal{L}_{CRE}$ denotes the cross-entropy loss. The terms $\lambda_1$ and $\lambda_2$ are hyperparameters that control the relative importance of the representation alignment loss $\mathcal{L}_{RA}$ and the JS divergence loss $\mathcal{L}_{JS}$ respectively.
\subsection{Analysis of Privacy and Additional Costs}
Notably, the computational and transmission overhead associated with sample statistics is negligible compared to the costs of model training and transmission. Additionally, attackers cannot reconstruct samples from these statistics or infer model data distribution on each client, thus enhancing privacy protection.

Furthermore, as shown in \cref{fig:upload_ratio&hyper}, our method achieves better performance than other baselines even with a minimal upload ratio $r$, further reducing the computational and transmission overhead as well as the risk of privacy leakage.
\section{Experiments}

\begin{table*}[ht]
\centering
\resizebox{0.99\textwidth}{!}{
\begin{tabular}{c|ccccc|ccccc|cccccc}
\toprule
\multicolumn{1}{c|}{} & \multicolumn{5}{c|}{\textbf{PACS}}                                                           & \multicolumn{5}{c|}{\textbf{OfficeHome}}                                                      & \multicolumn{5}{c}{\textbf{miniDomainNet}}                                         \\
\multicolumn{1}{c|}{} & P              & A              & C             & S              & \multicolumn{1}{c|}{Avg.} & A              & C              & P              & R              & \multicolumn{1}{c|}{Avg.} & C              & P              & R              & S              & Avg.           \\ \midrule
FedAvg                & 90.48          & 69.19          & 76.15         & 71.62          & 76.86                     & 62.30          & 49.51          & 76.12          & 77.55          & 66.37                     & 63.90          & 57.64          & 67.90          & 49.63          & 59.77          \\
FedProx               & 90.60          & 69.09          & 74.32         & 70.15          & 76.04                     & 63.08          & 50.13          & 76.28          & 77.65          & 66.78                     & 63.74          & 57.65          & 68.02          & 49.81          & 59.81          \\
FedADG                & 91.02          & 65.04          & 72.95         & 65.82          & 73.71                     & 60.82          & 47.93          & 71.93          & 75.58          & 64.06                     & 54.79          & 52.32          & 62.37          & 41.60          & 52.77          \\
GA                    & 92.81          & 66.50          & 76.19         & 69.56          & 76.26                     & 61.80          & 49.60          & 75.51          & 77.09          & 66.00                     & 63.80          & 57.37          & 67.62          & 49.83          & 59.66          \\
FedSR                 & 91.74          & 72.36          & 75.55         & 70.48          & 77.53                     & \textbf{64.19} & 50.29          & 76.39          & 77.65          & 67.13                     & 62.98          & 56.79          & 67.35          & 49.39          & 59.13          \\
FedIIR                & 91.26          & 71.73          & 77.94         & 71.32          & 78.06                     & 61.48          & 49.97          & 75.40          & 77.25          & 66.03                     & 63.90          & 57.11          & 67.90          & 49.33          & 59.56          \\
CCST                  & 90.16          & 76.37          & 76.11         & 78.58          & 80.30                     & 61.20          & 50.48          & 76.53          & 77.48          & 66.42                     & 62.75          & 56.54          & 65.32          & 47.93          & 58.14          \\
FedCCRL               & \textbf{93.83} & \textbf{79.15} & \textbf{77.9} & \textbf{78.95} & \textbf{82.46}            & 63.08          & \textbf{55.67} & \textbf{76.77} & \textbf{77.71} & \textbf{68.31}            & \textbf{66.33} & \textbf{59.68} & \textbf{68.03} & \textbf{54.50} & \textbf{62.14} \\ \bottomrule
\end{tabular}
}
\caption{Test accuracy on each dataset. In this set of experiments, we set the upload ratio of the statistics $r=0.1$. To ensure fairness, each algorithm is evaluated three times, and the final average is taken as the experimental result.}
\label{tab: main result}
\end{table*}
\begin{figure*}
    \centering
    \includegraphics[width=1\textwidth]{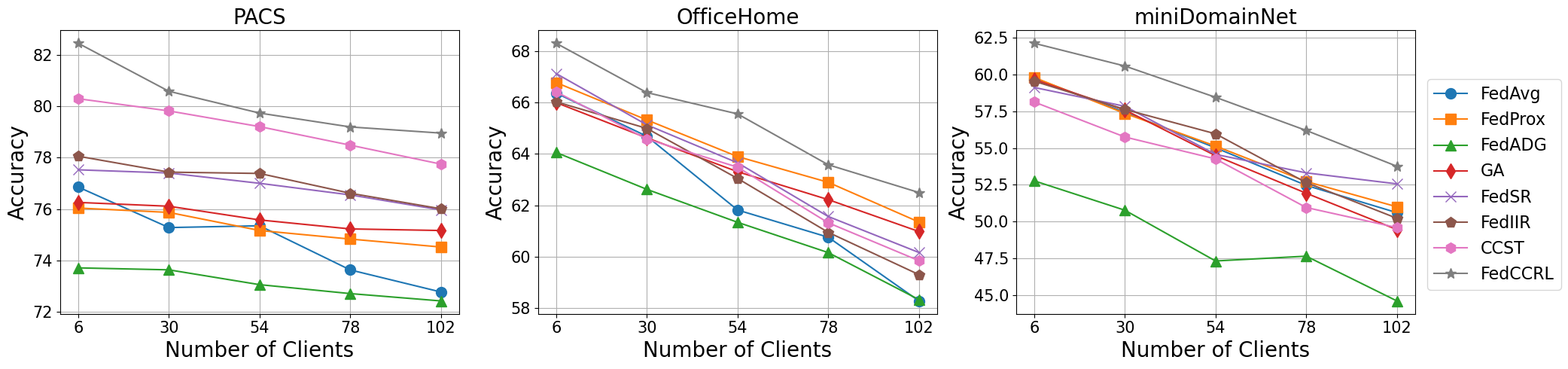}
    \caption{Average test accuracy (\%) versus the number of participating clients.}
    \label{fig:client_num}
\end{figure*}
\subsection{Dataset and Evaluation Protocol}
Our algorithm and baselines are evaluated on three widely used datasets in DG: PACS \cite{PACS}, OfficeHome \cite{officehome}, and miniDomainNet \cite{minidomainnet}. PACS comprises 9,991 samples across four domains: Art Painting, Cartoon, Photo, and Sketch, and is categorized into seven classes. OfficeHome contains 15,588 samples from four domains: Art, Clipart, Product, and Real World, and is divided into 65 classes. miniDomainNet features 140,006 samples from four domains: Clipart, Infograph, Painting, and Real, and encompasses a total of 126 classes.

We adopt the leave-one-domain-out evaluation protocol \cite{PACS}, where one domain is held out as the test set while the remaining domains are used for training. This process is repeated for each domain to ensure robust evaluation across varying distribution shifts.

\subsection{Baselines and Implementation Details}
In this experiment, we compare FedCCRL with several traditional FL frameworks, including FedAvg \cite{fedavg} and FedProx \cite{fedprox}, as well as state-of-the-art FDG approaches such as FedADG \cite{fedadg}, FedSR \cite{fedsr}, FedIIR \cite{fediir}, GA \cite{ga}, and CCST \cite{ccst}.

Following the data partition scheme outlined in \cref{sec: preliminary}, we evaluate FedCCRL and the baselines across scenarios with different numbers of clients. We utilize MobileNetV3-Large \cite{mobile_net_v3} as the backbone network, where the final fully connected layer acts as the classifier $g$, and the preceding layers serve as the representation encoder $h$.

For training, we set the number of communication rounds to 10, and the number of local training epochs is fixed at 3 for all methods. The Adam optimizer \cite{adam} is employed with an initial learning rate of 0.001. Furthermore, CCDT is configured with the parameter \( \alpha = 0.1 \), while for DIFP, the parameter \( \beta \) is set to 1.0. For supervised contrastive loss, we set the temperature parameter $\tau$ to 0.1. Input samples from the PACS and OfficeHome datasets are resized to $224\times224$, whereas those from miniDomainNet are resized to $128\times128$. To manage the learning rate, we adopt a cosine learning rate scheduler, which facilitates a smooth decay of the learning rate throughout the training process.

To guarantee the reliability of our proposed algorithms, we run each experiment three times independently and report the averaged results. 

\subsection{Comparison with Baselines}
We evaluate FedCCRL and other baselines under scenarios with different numbers of clients. \Cref{tab: main result} specifically shows the performance of each algorithm when the total number of clients is set to 6. In this table, we use the first letter of each domain to represent the domain, and each letter column corresponds to the performance when that domain is used as the target domain. The "Avg." column provides the average results across the four leave-one-domain-out experiments. Additionally, \cref{fig:client_num} presents the performance of each algorithm under varying client numbers, where the results are averaged across the four leave-one-domain-out experiments. In these experiments, the upload ratio for FedCCRL is set to 0.1.

From \cref{tab: main result}, we have the following observations:
\begin{itemize}
    \item FedCCRL outperforms all baseline methods on all datasets. Specifically, FedCCRL achieves the best average accuracy on all datasets. In addition, FedCCRL attains top scores on each test domain of both PACS and miniDomainNet datasets.
    \item Most baselines specifically designed for FDG fail to achieve consistent performance across different datasets. In some cases, their performance is even inferior to traditional federated learning algorithms such as FedAvg and FedProx.
\end{itemize}

Additionally, as shown in \cref{fig:client_num}, although the performance of all baselines significantly declines as the number of participating clients increases, FedCCRL consistently outperforms the baselines across both small- and large-scale client scenarios, which illustrate that FedCCRL is robust to the size of the FL system.

\subsection{Ablation Study}
\begin{table}
\centering
\resizebox{0.99\linewidth}{!}{
\begin{tabular}{c|c|c|c|ccc}
\toprule
CCDT                                    & DIFP                                      & $\mathcal{L}_{RA}$                  & $\mathcal{L}_{JS}$                 & \textbf{PACS} & \textbf{OfficeHome} & \textbf{miniDomainNet} \\ \midrule
\multirow{4}{*}{\ding{51}} & \multirow{4}{*}{}                           &                                     &                                     & 79.92         & 66.58               & 59.57                  \\
                                            &                                             & \ding{51}          &                                     & 80.95         & 66.93               & 61.43                  \\
                                            &                                             &                                     & \ding{51}          & 80.52         & 67.34               & 60.82                  \\
                                            &                                             & \ding{51}          & \ding{51}          & 81.11         & 67.58               & 61.83                  \\ \midrule
\multirow{4}{*}{}                           & \multirow{4}{*}{\ding{51}} &                                     &                                     & 79.32         & 66.84               & 59.64                  \\
                                            &                                             & \ding{51}          &                                     & 80.62         & 67.29               & 60.94                  \\
                                            &                                             &                                     & \ding{51}          & 80.38         & 67.68               & 61.53                  \\
                                            &                                             & \ding{51}          & \ding{51}          & 80.97         & 67.92               & 61.94                  \\ \midrule
\multirow{4}{*}{\ding{51}} & \multirow{4}{*}{\ding{51}} &                                     &                                     & 80.48         & 67.29               & 60.40                  \\
                                            &                                             & \ding{51}          &                                     & 81.18         & 67.81               & 61.47                  \\
                                            &                                             &                                     & \ding{51}          & 80.90         & 68.08               & 61.49                  \\
                                            &                                             & \textbf{\ding{51}} & \textbf{\ding{51}} & \textbf{82.46}         & \textbf{68.31}               & \textbf{62.14}                  \\ \bottomrule
\end{tabular}
}
\caption{Ablation study results. The total number of clients is set to 6 and the upload ratio is set to 0.1.}
\label{tab: ablation study}
\end{table}

We investigate the effects of CCDT, DIFP, representation alignment regularizer $\mathcal{L}_{RA}$, and JS divergence loss $\mathcal{L}_{JS}$ on FedCCRL across the PACS, OfficeHome, and miniDomainNet datasets. The results are presented in \cref{tab: ablation study}, from which we can draw the following conclusions:
\begin{itemize}
    \item All four components significantly enhance the model's generalization ability. The utilization of representation alignment loss or JS divergence loss independently demonstrates superior performance compared to employing cross-entropy loss in isolation, while their combined application yields the most optimal results. Additionally, combining CCDT and DIFP  improves performance compared to using either method alone.
    \item Even without presentation alignment or prediction alignment, FedCCRL achieved accuracies of 80.48\%, 67.28\%, and 60.4\% on the PACS, OfficeHome, and miniDomainNet datasets, respectively, outperforming all other baselines.
\end{itemize}
\begin{figure*}
    \centering
    \includegraphics[width=\textwidth]{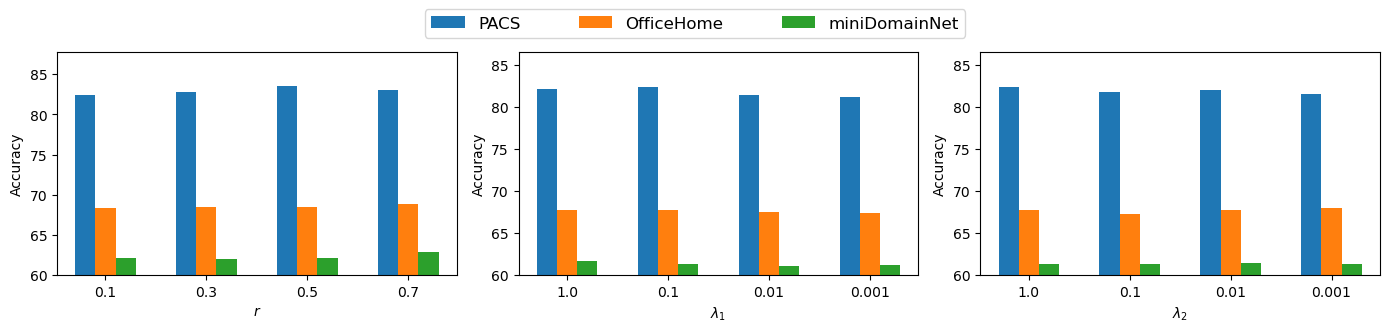}
    \caption{Effect of $r$ and $\lambda_1$, $\lambda_2$
  on accuracy across datasets}
    \label{fig:upload_ratio&hyper}
\end{figure*}
\begin{figure*}
    \centering
    \includegraphics[width=\textwidth]{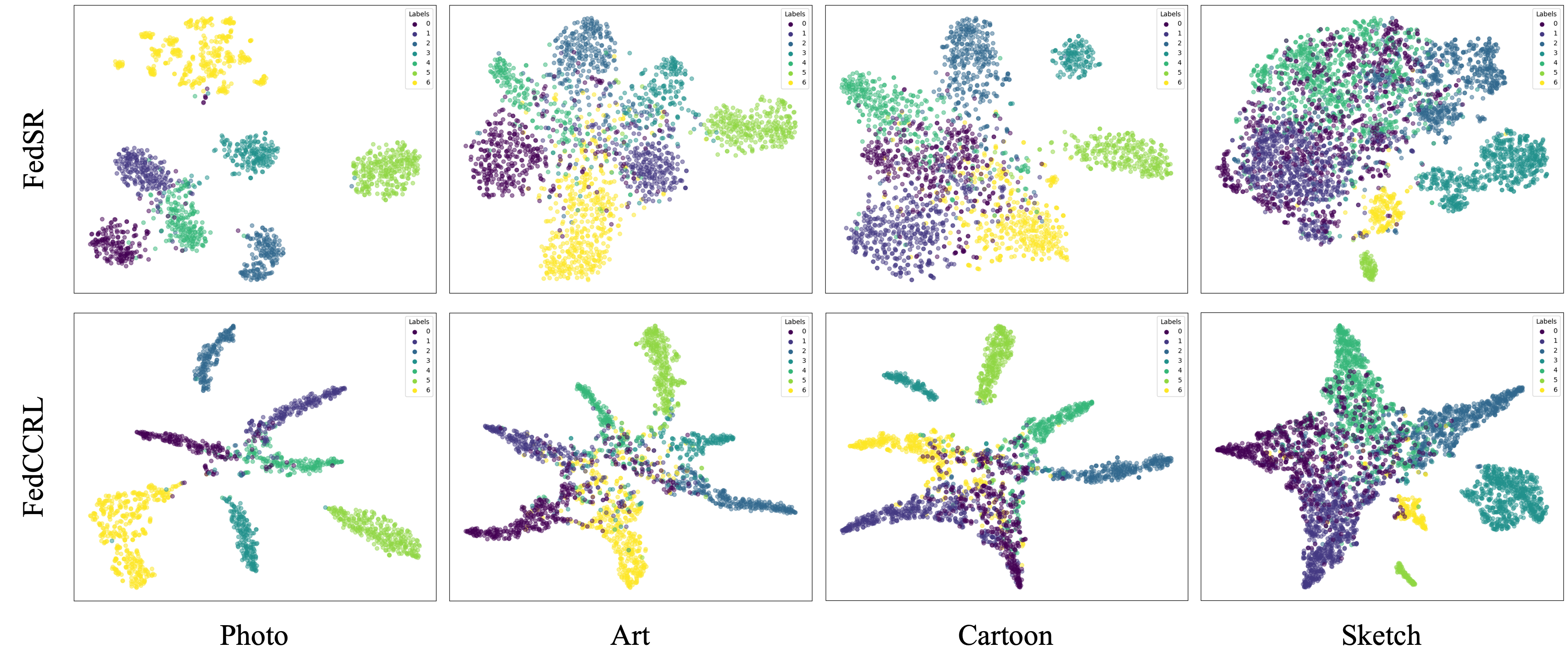}
    \caption{t-SNE visualization of representation distribution of each test domain on PACS dataset. Different colors denote different labels.}
    \label{fig:tsne_overall_distribution}
\end{figure*}

\subsection{Impact of Upload Ratio and Hyperparameters}
    


We evaluated the performance of FedCCRL under different upload ratios and hyperparameter settings, and the results are presented in \cref{fig:upload_ratio&hyper}. When evaluating the effect of $\lambda_1$, $\lambda_2$ is fixed at 1. Similarly, when assessing the impact of $\lambda_2$, $\lambda_1$ is set to 0.1.  

The evaluation results in \cref{fig:upload_ratio&hyper} demonstrate that both the upload ratio \( r \) and the hyperparameters \( \lambda_1 \) and \( \lambda_2 \) have minimal impact on the overall performance across the PACS, OfficeHome, and miniDomainNet datasets. Therefore, it can be concluded that FedCCRL is robust to variations in these parameters and outperforms other baselines even with only a very small amount of upload of statistics.
\subsection{Visualization of Representation Distribution and Model Attention}
To better illustrate the effect of FedCCRL on the model, we present the t-SNE visualization of representation distribution in \cref{fig:tsne_overall_distribution}, and Grad-CAM visualization of the model's concentration, provided as \cref{fig:grad-cam} in the supplementary material.

\Cref{fig:tsne_overall_distribution} demonstrates that FedCCRL achieves
more distinct and compact groupings of representations compared to FedSR, particularly in the Photo and Art domains, indicating that FedCCRL is effective in enabling model to learn domain-invariant features across different domains.
Additionally, \cref{fig:grad-cam} reveals that the model trained with FedCCRL focuses more effectively on domain-invariant features, such as the giraffe’s neck, the elephant’s face and human heads. This suggests that FedCCRL enhances the model’s ability to concentrate on relevant, discriminative regions in the images, leading to improved generalization across domains.
\section{Conclusion}
This paper introduces FedCCRL, a lightweight FDG framework that significantly enhances model's generalization ability while upholding privacy and incurring minimal computational and communication costs. By employing a lightweight cross-client feature extension module, FedCCRL effectively mitigates the challenges posed by limited local data quantity and domain diversity. Moreover, FedCCRL further enables the model to capture domain-invariant features with representation and prediction dual-stage alignment. 
Experimental results on benchmark datasets—PACS, OfficeHome, and miniDomainNet—demonstrate that FedCCRL outperforms existing state-of-the-art approaches across both small- and large-scale client scenarios. 
{
    \small
    \bibliographystyle{ieeenat_fullname}

}

\clearpage
\setcounter{page}{1}
\setcounter{section}{0}
\maketitlesupplementary
\section{Process of AugMix}
%
AugMix \cite{augmix} consists of multiple augmentation chains, each of which is composed of a series of diverse data augmentation operations. The diversity of augmented samples enhances the model's robustness by exposing the model to a wider variety of data distributions during training.
\begin{algorithm}
    \caption{AugMix}
    \label{alg: augmix}
    \textbf{Input:} A batch of samples $X$, Operations base $\mathcal{O}$, $\beta \in (0,\infty)$\\
    \textbf{Output:} A batch of augmented samples $X_{augmix}$
    \begin{algorithmic}[1]
    \State $\hat{X}=[ \ ]$
    \For{$x$ in $X$}
        \State Fill $x_{\text{aug}}$ with zeros
        \State Uniformly sample $k\sim\{1,2,3\}$
        \State Sample $(w_1, w_2, \ldots, w_k) \sim \text{Dirichlet}(\beta, \beta, \ldots, \beta)$
        \For{$i = 1, \ldots, k$}
            \State Sample operations $\text{op}_1, \text{op}_2, \text{op}_3 \sim \mathcal{O}$
            \State $\text{op}_{12} = \text{op}_2 \circ \text{op}_1$, $\text{op}_{123} = \text{op}_3 \circ \text{op}_2 \circ \text{op}_1$
            \State Uniformly sample $\text{chain} \sim \{\text{op}_1, \text{op}_{12}, \text{op}_{123}\}$
            \State $x_{\text{aug}} \mathrel{+}= w_i \cdot \text{chain}(x)$ \hfill \Comment{Element-wise Addition}
        \EndFor
        \State Sample weight $m \sim \text{Beta}(\beta, \beta)$
        \State $\hat{x} = m x + (1 - m) x_{\text{aug}}$
        \State $\hat{X}.\text{append}(\hat{x})$
    \EndFor
    \State $X_{augmix}=\text{cat}(\hat{X})$
    \end{algorithmic}
\end{algorithm}

\section{Visualization of Model's Attention}

\begin{figure}[H]
    \centering
    \includegraphics[width=1\linewidth]{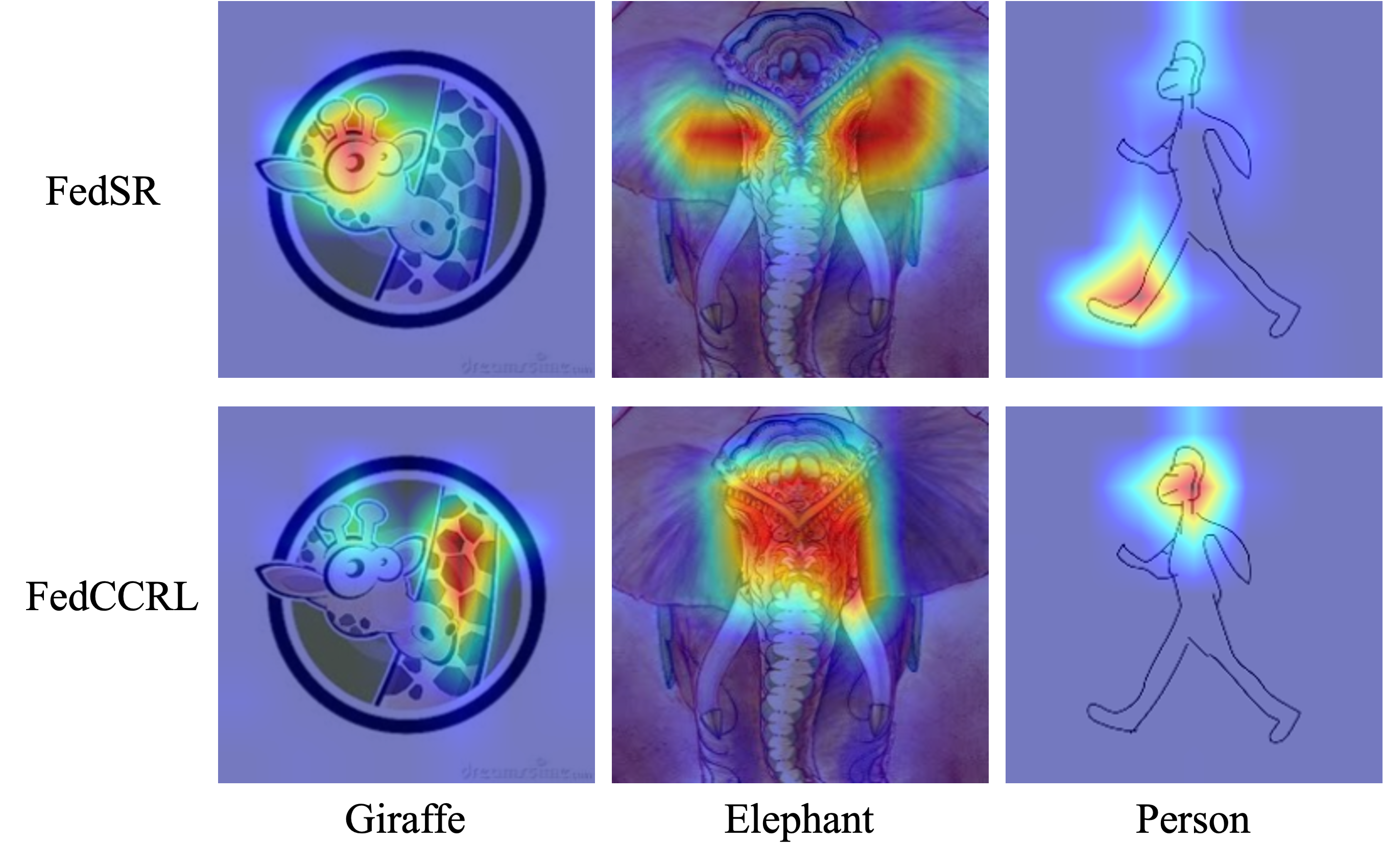}
    \caption{Grad-CAM visualization on the PACS dataset.}
    \label{fig:grad-cam}
\end{figure}
We visualize the model's attention by Grad-CAM algorithm in Figure \ref{fig:grad-cam}. The results demonstrate that the model trained with FedCCRL effectively identifies features that are semantically aligned with the corresponding labels, such as the giraffe's neck, the elephant's face, and the human head.


\end{document}